\pdfoutput=1

\documentclass[11pt]{article}

\usepackage[final]{acl}
\usepackage{booktabs}
\usepackage{times}
\usepackage{latexsym}
\usepackage{multirow} 
\usepackage{array}
\usepackage{pifont}   
\usepackage{arydshln}
\usepackage[T1]{fontenc}
\newcommand{\figref}[1]{Fig. \ref{#1}}
\newcommand{\tabref}[1]{Table \ref{#1}}
\newcommand{\secref}[1]{Section \ref{#1}}

\usepackage[utf8]{inputenc}
\usepackage{tcolorbox}
\usepackage{bm}
\usepackage{amsfonts}

\usepackage{microtype}
\usepackage{framed}
\usepackage{mdframed}
\usepackage{inconsolata}
\usepackage{amsmath} 
\usepackage{graphicx}
\usepackage{fontawesome}
\usepackage{xcolor}
\usepackage[misc]{ifsym}
\definecolor{kellygreen}{rgb}{0.3, 0.73, 0.09}
\definecolor{alizarin}{rgb}{0.82, 0.1, 0.26}

\title{\textsc{EifBench}: Extremely Complex Instruction Following Benchmark\\ for Large Language Models}

\author{
Tao Zou, Xinghua Zhang, Haiyang Yu, Minzheng Wang, Fei Huang, Yongbin Li \thanks{*Corresponding author} \\
    \normalsize{Tongyi Lab, Alibaba Group} \\
     {\small \texttt{\{qingdie.zt, zhangxinghua.zxh, yifei.yhy, wangminzheng.wmz, f.huang, shuide.lyb\}@alibaba-inc.com}}
}

\begin{document}
\maketitle
\begin{abstract}
With the development and widespread application of large language models (LLMs), the new paradigm of ``{\it Model as Product}'' is rapidly evolving, and demands higher capabilities to address complex user needs, often requiring precise workflow execution which involves the accurate understanding of multiple tasks.
However, existing benchmarks focusing on single-task environments with limited constraints lack the complexity required to fully reflect real-world scenarios. 
To bridge this gap, we present the \textbf{E}xtremely Complex \textbf{I}nstruction \textbf{F}ollowing \textbf{Bench}mark (\textbf{\textsc{EifBench}}), meticulously crafted to facilitate a more realistic and robust evaluation of LLMs. 
\textsc{EifBench} not only includes multi-task scenarios that enable comprehensive assessment across diverse task types concurrently, but also integrates a variety of constraints, replicating complex operational environments.
Furthermore, we propose the \textbf{Seg}ment \textbf{P}olicy \textbf{O}ptimization (\textbf{SegPO}) algorithm to enhance the LLM's ability to accurately fulfill multi-task workflow.
Evaluations on \textsc{EifBench} have unveiled considerable performance discrepancies in existing LLMs when challenged with these extremely complex instructions. This finding underscores the necessity for ongoing optimization to navigate the intricate challenges posed by LLM applications.
\end{abstract}

\section{Introduction}
\label{section-1}
The advent of large language models (LLMs) has transformed real-world applications by improving models' ability to comprehend a diverse range of human instructions, from simple conversations to complex problem solving \cite{DBLP:conf/iclr/SanhWRBSACSRDBX22, DBLP:conf/nips/DuboisLTZGBGLH23, DBLP:journals/corr/abs-2409-14195}. Thus, instructions have become central to effective human-machine interaction in this new landscape \cite{DBLP:conf/emnlp/ZhongLZK21, DBLP:conf/acl/MishraKBH22, DBLP:conf/chi/GaoGCLPM24}, especially the paradigm of ``{\it Model as Product}'' has deeply entered the collective consciousness where LLM agents need to accurately complete a series of tasks to meet user demands~\cite{xiong2025rag,pmlr-v235-hu24s,alakuijala2025memento}. However, as user demands grow more sophisticated, traditional benchmarks \cite{DBLP:conf/naacl/ZhongCGLLWSCD24, DBLP:journals/corr/abs-2306-04757}, which focus on specific tasks, are insufficient to evaluate models' comprehensive ability to handle multifaceted instructions. This shortfall underscores the need for innovative evaluation frameworks capable of accurately assessing how models understand and execute complex instructions \cite{DBLP:journals/corr/abs-2311-07911, DBLP:conf/acl/WangKMLSKH23, DBLP:conf/iclr/XuSZG0FTLJ24}.

\begin{figure}[!t]
    \centering
    \includegraphics[width=1\columnwidth]{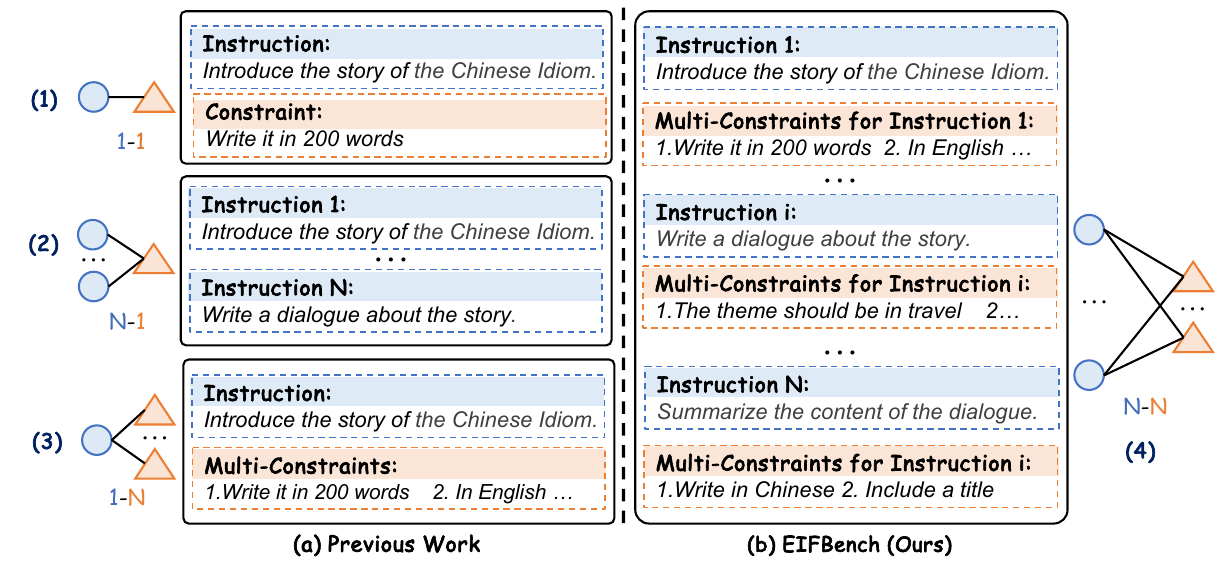}
\caption{Existing benchmarks, represented on the left, either focus on completing a single instruction or handling multiple instructions with only one constraint each. In contrast, \textsc{EifBench} presents a multi-instruction, multi-constraint benchmark, designed to more closely align with real-world complexities and demands.}


    \label{fig:back}
\end{figure}

To evaluate the instruction following abilities of LLMs, several benchmarks \cite{DBLP:journals/corr/abs-2311-07911, DBLP:conf/acl/QinSHYCWW00Y24, DBLP:conf/acl/LiZQLLWLYMZZLZM24} have been proposed, which can be categorized into three main types as shown in Fig.~\ref{fig:back}: (1) {\it \textbf{Single-Instruction Single-Constraint}} benchmarks, such as IFEval \cite{DBLP:journals/corr/abs-2311-07911} and \textsc{InFoBench} \cite{DBLP:conf/acl/QinSHYCWW00Y24}, focus on tasks governed by a single constraint, providing insights into basic instruction following abilities. (2) {\it \textbf{Single-Instruction Multi-Constraint}} benchmarks, like CFBench \cite{DBLP:conf/aaai/HeZHCXHZLX24}, evaluate how models handle a single instruction with multiple constraints across content, numerical, and other dimensions simultaneously. (3) {{\it \textbf{Multi-Instruction Single-Constraint}} scenarios, such as those explored by SIFo \cite{DBLP:conf/emnlp/ChenLQEMBR24}, test models' adherence to sequences of instructions, assessing their adaptability and versatility while maintaining focus on a single constraint. Nonetheless, research still lacks in addressing {\textbf{multi-instruction multi-constraint}} scenarios, which more accurately reflect real-world complexities, especially in the era of LLMs serving as agents with workflow execution involving multiple tasks.

\begin{table*}[t]
\renewcommand\arraystretch{1.1}
    \centering
    \resizebox{\textwidth}{!}{%
    \begin{tabular}{lccccc}
        \toprule
        \textbf{Benchmark} & \textbf{Multi-Constraint} & \textbf{Multi-Instruction} & \textbf{Multi-Type} & \textbf{Average Constraint} & \textbf{Average Instruction} \\
        \midrule
        \textbf{CIF-Bench} \cite{DBLP:conf/acl/LiZQLLWLYMZZLZM24} & \textcolor{red}{\ding{55}} & \textcolor{red}{\ding{55}} & \textcolor{red}{\ding{55}} &1.00  &1.00  \\
        \textbf{FollowBench} \cite{DBLP:conf/acl/Jiang0ZZLMS00W24} & \textcolor{kellygreen}{\ding{51}} & \textcolor{red}{\ding{55}} & \textcolor{red}{\ding{55}} &3.00  &1.00  \\
        \textbf{ComplexBench} \cite{DBLP:journals/corr/abs-2407-03978} & \textcolor{kellygreen}{\ding{51}} & \textcolor{red}{\ding{55}} & \textcolor{red}{\ding{55}} &4.19  &1.00  \\
        \textbf{CFBench} \cite{DBLP:conf/aaai/HeZHCXHZLX24} & \textcolor{kellygreen}{\ding{51}} & \textcolor{red}{\ding{55}} & \textcolor{red}{\ding{55}} &4.24  &1.00  \\
        \textbf{SIFo} \cite{DBLP:conf/emnlp/ChenLQEMBR24} & \textcolor{red}{\ding{55}} & \textcolor{kellygreen}{\ding{51}}  & \textcolor{red}{\ding{55}} &1.00  &4.17  \\ \hdashline
        \textbf{\textsc{EiFbench} (Ours)} & \textcolor{kellygreen}{\ding{51}} & \textcolor{kellygreen}{\ding{51}}  & \textcolor{kellygreen}{\ding{51}} & \textbf{74.01}  & \textbf{8.24}  \\
        \bottomrule
    \end{tabular}%
    }
    \caption{\textsc{Eifbench} encompasses multi-instruction multi-constraint samples across multiple data types. ``Multi-type'' refers to the inclusion of data from various formats, such as plain text, dialogue, and multi-party dialogue, highlighting diverse communication styles and structures.}
\end{table*}

Multi-instruction multi-constraint (MIMC) scenarios are ubiquitous in real-world applications, such as workflow automation \cite{DBLP:conf/nips/ZhangLG22, taylor2023accelerated} and healthcare scheduling \cite{DBLP:journals/eaai/BakhshandehA24, DBLP:conf/isr2/LiJSZY21}. For example, in cloud-based workflow automation, orchestrating computational tasks such as data preprocessing, model inference, and report generation requires balancing resource allocation, execution time, and task dependencies \cite{xiong2016deadline}. However, existing LLMs struggle with such complexity, with performance dropping by over 30\% with over 5 constraints \cite{DBLP:conf/aaai/HeZHCXHZLX24}. Bridging this gap necessitates benchmarks that mirror real-world MIMC dynamics, integrating both task interdependence and constraint scalability to foster robust and adaptable LLMs.

In response to these challenges, we introduce the \textbf{E}xtremely Complex \textbf{I}nstruction \textbf{F}ollowing \textbf{Bench}mark (\textbf{\textsc{EifBench}}), specifically designed to address the shortcomings of current benchmarks by providing a comprehensive framework that mirrors the complexities of real-world task environments. As shown in \figref{fig:back}, \textsc{EifBench} is unique in its inclusion of multi-task scenarios, drawn from diverse sources and integrated with multifaceted constraints\footnote{In this work, plain text datasets refer to non-conversational plain text datasets.}. This design allows for an in-depth assessment of a model's ability to manage complex demands. 
In addition, we introduce the \textbf{Segment Policy Optimization (SegPO)} algorithm, which features advantage estimation for outputs corresponding to each instruction within multi-instruction inputs. 
The main contributions of this paper are summarized as follows:
\begin{itemize}
    
    \item We first develop the extremely complex instruction following benchmark (\textsc{EifBench}), simulating real-world applications with multiple instructions and constraints.

    \item We propose the segment policy optimization (SegPO) algorithm by calculating the advantages separately for each output that responds to the corresponding instruction within the input, encouraging more nuanced feedback in following multiple instructions.
    
    \item We conduct a detailed analysis of 20 LLMs, encompassing both open-source and closed-source models, uncovering their limitations in processing complex instructions and pinpointing areas for enhancement to better adapt to real-world complex scenarios. The SegPO algorithm demonstrates significant improvements, achieving increases of 14.85\% compared to the base LLM and 3.40\% compared to GRPO models on \textsc{EifBench}, respectively.
\end{itemize}
\section{\textsc{EifBench}}
\label{section-2}
\subsection{Task and Constraint Taxonomy}
To thoroughly assess the capability of large language models (LLMs) in adhering to complex instructions, we introduce an exceptionally challenging instruction following benchmark. Specifically, we categorize both tasks and constraints to structure the evaluation. For tasks, we identify and compile 8 types of tasks based on traditional NLP tasks. Regarding constraints, we establish a two-level hierarchical taxonomy for the organization.

\subsubsection{Task Categories}
\label{tasks}
In line with instruction following existing works \cite{DBLP:journals/corr/abs-2408-01122, DBLP:conf/acl/LiZQLLWLYMZZLZM24}, we categorize the tasks in \textsc{EifBench} into eight primary types. \footnote{In this work, following an instruction refers to completing one specific task and producing the corresponding response.} These categories provide a comprehensive framework for systematically evaluating model performance across diverse task settings. The distribution of these task categories is shown in \figref{fig:task_dis}.

\begin{figure}[!t]
    \centering
    \includegraphics[width=0.9\columnwidth]{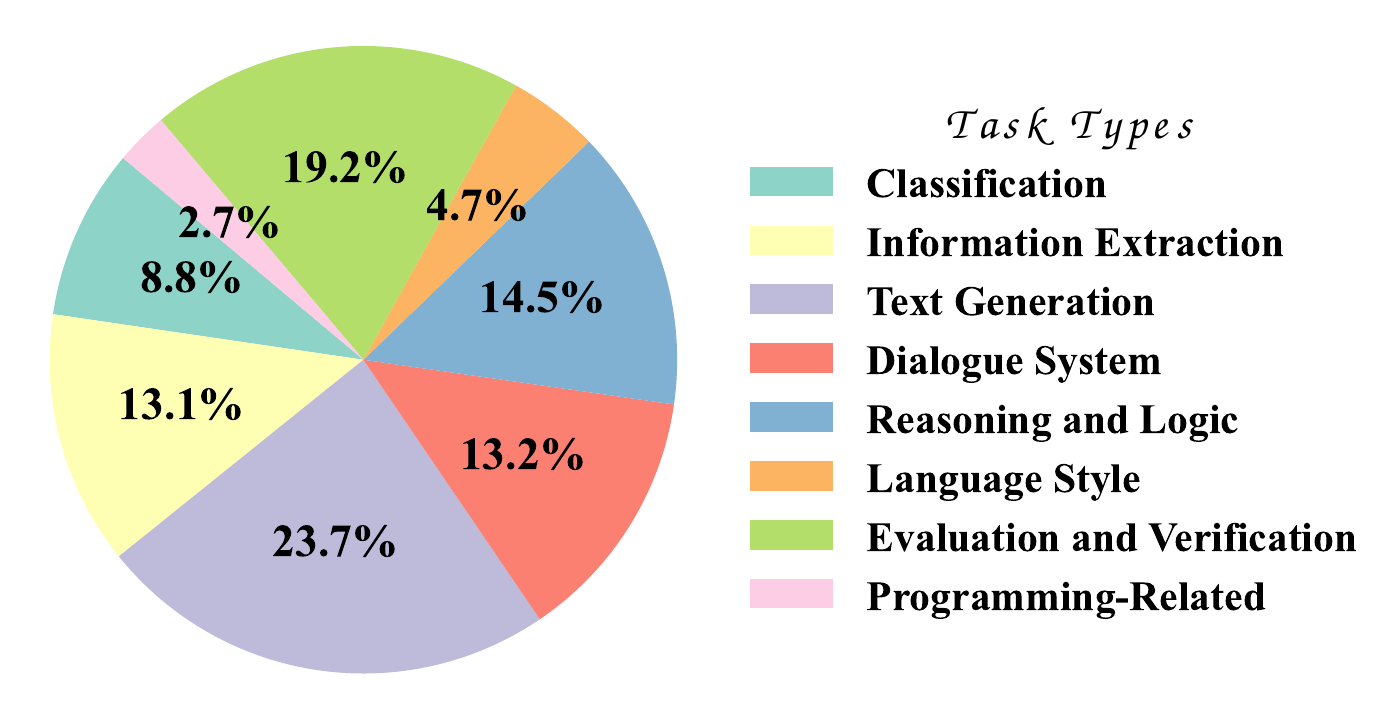}
    \caption{Task type distribution in \textsc{EifBench}.}
    \label{fig:task_dis}
\end{figure}


\noindent\textbf{Classification} involves sentiment analysis, text and toxic content classification, empathy detection, and social norm judgment.

\noindent\textbf{Information Extraction} focuses on extracting key information such as named entity recognition, keyword annotation, and entity relationships.

\noindent\textbf{Text Generation} tasks cover creative and practical outputs, including story generation, text expansion, and headline content generation.

\noindent\textbf{Dialogue System} tasks are designed for developing interactive agents through dialogue generation, intent recognition, and state information tracking.

\noindent\textbf{Reasoning and Logic} tasks require logical inference and critical thinking, including commonsense and multi-hop reasoning question answering.

\noindent\textbf{Language Style}  tasks involve style manipulation and analysis, such as style transfer, sarcasm detection, and dialect variation recognition.

\noindent\textbf{Evaluation and Verification} tasks concentrate on verifying information and assessing text quality, including fact consistency verification.

\noindent\textbf{Programming-Related} tasks evaluate programming understanding through code generation, debugging, and explanation capabilities.

In addition, tasks are structured into distinct modes: parallel for simultaneous dimension consideration, serial for chain dependencies, conditional for adaptability to varying conditions, and nested for hierarchical structures. These categories provide a systematic evaluation of model capabilities in the benchmark.

\subsubsection{Constraint Categories}
\label{constraint}
Following established research on instruction following \cite{DBLP:journals/corr/abs-2411-06208}, we have developed a comprehensive constraint system for \textsc{EifBench}. This system categorizes constraints into four primary types: Content Constraints, Situation Constraints, Style Constraints, and Format Constraints. These categories provide a structured framework to systematically evaluate the capabilities of language models across a wide range of instructional scenarios. The distribution is shown in \figref{fig:constraint_dis}. Detailed descriptions of the specific constraint dimensions within each category are provided in Appendix \ref{app:con}.

\begin{figure}[!t]
    \centering
    \includegraphics[width=1.0\columnwidth]{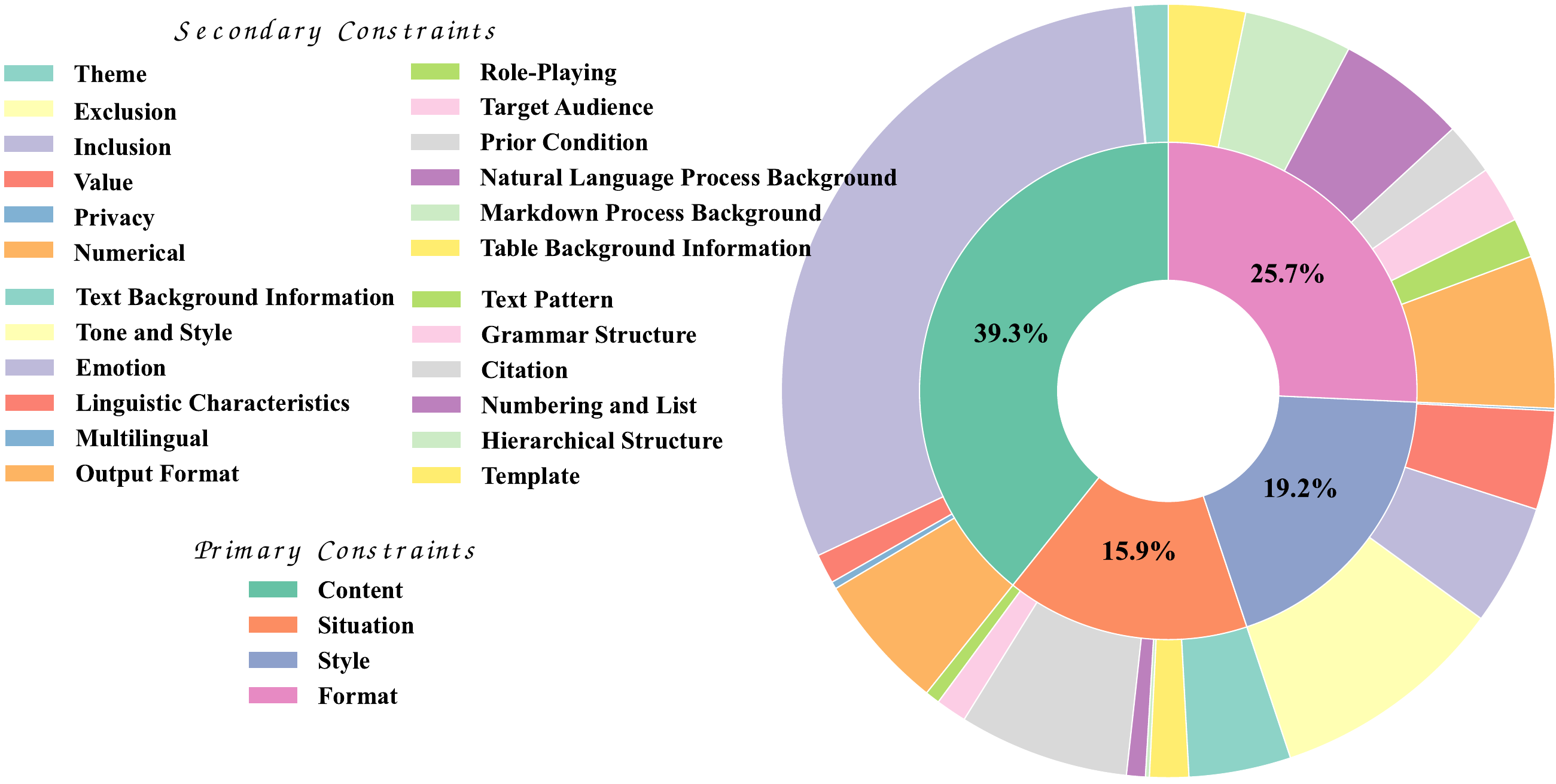}
    \caption{Constraint type distribution in \textsc{EifBench}.}
    \label{fig:constraint_dis}
\end{figure}

    \noindent \textbf{Content Constraints}. These ensure the text follows specific thematic topics, inclusion/exclusion criteria, values, tone, style, privacy considerations, and numerical precision.

    \noindent \textbf{Situation Constraints}. These emphasize contextual elements like audience specifications, preconditions, and incorporate various knowledge and background information formats.

    \noindent \textbf{Style Constraints}. These govern tone, emotion, style, and multilingual features to suit the required stylistic and emotional text aspects.

    \noindent \textbf{Format Constraints}. These ensure adherence to essential structural requirements such as output formats, text patterns, grammar, accurate sentence structure, and hierarchical organization.








\subsection{Construction Workflow}
\label{section-3}
\begin{figure*}[!htbp]
    \centering
    \includegraphics[width=2.0\columnwidth]{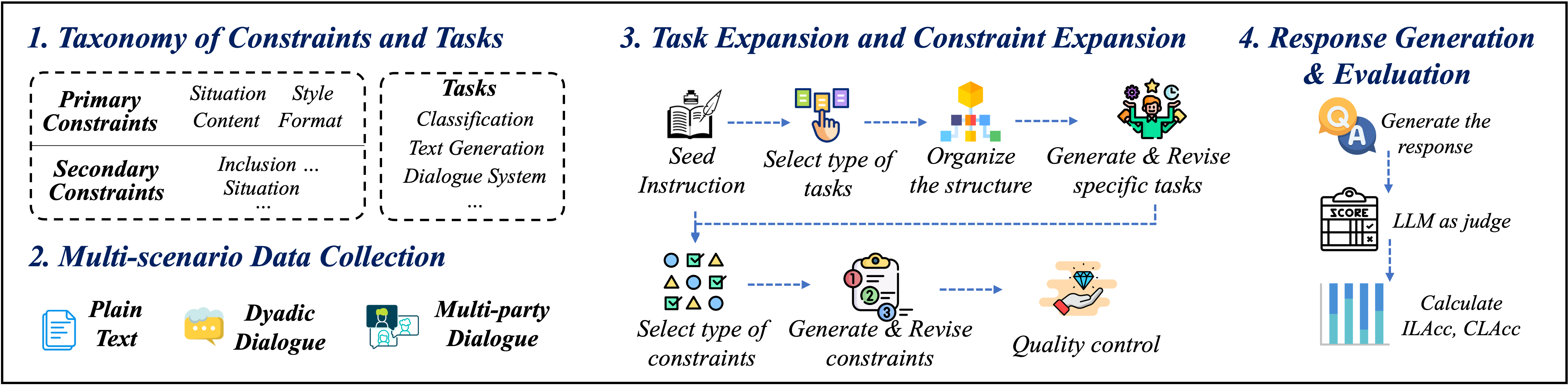}
    \caption{Pipeline for constructing the benchmark.}
    \label{fig:framework}
\end{figure*}

The overall construction process includes several key stages: 1) Taxonomy of Constraints and Tasks, 2) Multi-scenario Data Collection, 3) Task Expansion, 4) Constraint Expansion, 5) Quality Control, and 6) Response Generation \& Evaluation.

\textbf{1) Taxonomy of Constraints and Tasks}. We establish two taxonomies for constraints and tasks, as presented in \secref{section-2}.


\textbf{2) Multi-scenario Data Collection}. Our collection process involves three types of datasets: plain text, dyadic dialogue, and multi-party dialogue. Plain text samples are drawn from existing works \cite{DBLP:journals/corr/abs-2407-03978, DBLP:conf/acl/LiZQLLWLYMZZLZM24}. For dyadic dialogues, we first perform data cleaning and noise reduction on the collected real-life interactions. Guided by the methodology in \citet{DBLP:conf/acl/00010CX00ML25}, we employ Large Language Models (LLMs) to condense the conversations while ensuring the preservation of key information. Multi-party dialogue data is synthesized with LLMs, crafting diverse scenarios and participant numbers. Specific prompts guide LLMs to produce varied and representative dialogue content, enhancing the depth and applicability of the dataset.

\textbf{3) Task Expansion}. Tasks are expanded into series in the plain text scenario (see \secref{tasks}). Using LLMs, we develop complex task sets with dependencies and parallelism. We also conduct rigorous quality assessments, removing redundant, infeasible, and contradictory tasks, thus ensuring the quality and consistency of the generated data. In dyadic and multi-party dialogue scenarios, we directly generate multiple new tasks, ensuring each reflects the complexity of real-world interactions.



\textbf{4) Constraint Expansion}. In the constraint expansion process, we refine simple instructions using a predefined taxonomy (see \secref{constraint}). Utilizing LLMs, complexity is incrementally added, ensuring tasks encompass a broad spectrum of requirements and constraints. This iterative review targets and clarifies ambiguous semantics to ensure constraints are objectively evaluated and quantified. This method not only adds complexity and challenge but also enhances the realism and comprehensiveness of the data generated.

\textbf{5) Quality Assessment}. Our quality assessment covers instruction-level and constraint-level validation. For instruction-level validation, we ensure logical consistency and feasibility for LLMs, removing contradictory, redundant, or infeasible tasks while maintaining a diverse, moderate difficulty task set of 6 to 12 instructions. In constraint-level validation, constraints are iteratively refined using predefined taxonomies, ensuring they are objectively quantified and within model capabilities, addressing any ambiguity or infeasibility.


\textbf{6) Response Generation \& Evaluation}. First, using the instruction data, we employ various language models to generate the corresponding outputs. To verify their compliance, we then prompt large language models to assess each constraint satisfaction for the outputs, generating a binary outcome (0/1) that indicates whether the generated output satisfies the respective constraints.


\begin{table}[t]
\centering
\small
\renewcommand\arraystretch{1.1}
\tabcolsep=0.12cm
\begin{tabular}{lcccc}
\toprule
Category & \#N & Min. & Max. & Avg. \\
\midrule
Plain Text & 450 & 41 & 107 & 73.27 \\
Dyadic Dialogue & 450 & 47 & 107 & 73.38 \\ 
Multi-party Dialogue & 100 & 63 & 116 & 80.26 \\ 
\bottomrule
\end{tabular}
\caption{Statistics of \textsc{EiFbench}. \#N denotes data instances; Min., Max., and Avg. mean the minimum, maximum, and average number of constraints per instance.}
\label{table:statistics}
\end{table}

As shown in \tabref{table:statistics}, \textsc{EifBench} comprises 1,000  instances. Across three subsets, the minimum, maximum, and average numbers of constraints per instance are reported. \figref{fig:constraints} and \tabref{tab:sub-tasks} illustrate the distribution of constraint numbers and instruction numbers within \textsc{EifBench}.

\begin{figure}[!t]
    \centering
    \includegraphics[width=1.0\columnwidth]{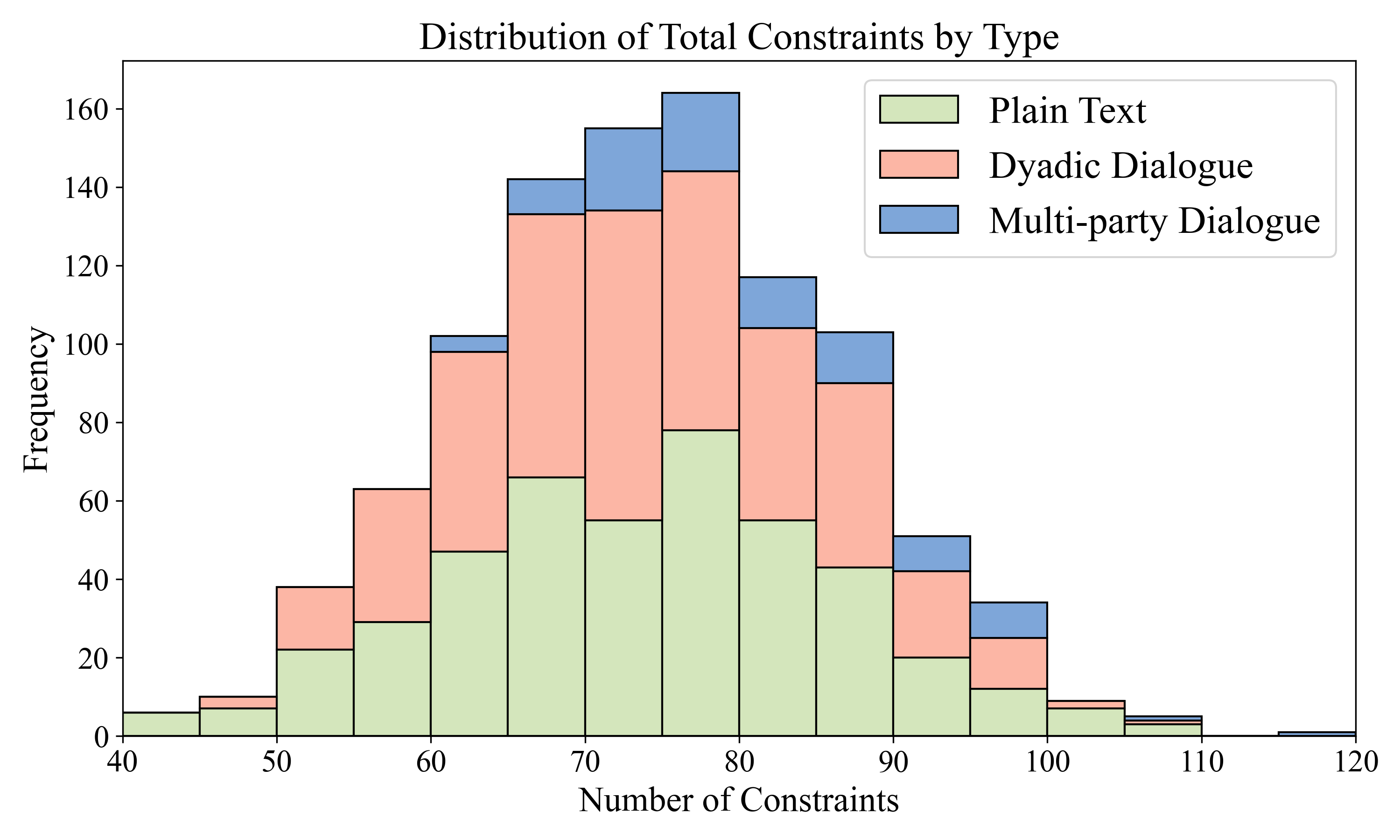}
    \caption{Distributions of total constraints for different text categories.}
    \label{fig:constraints}
\end{figure}

\begin{table}[!t]
\centering
\small
\renewcommand\arraystretch{1.1}
\tabcolsep=0.075cm
\begin{tabular}{lccccccc}
\toprule
Scenario & 6 & 7 & 8 & 9 & 10 & 11 & 12 \\
\midrule
Plain Text & 15 & 76 & 136 & 139 & 76 & 7 & 1 \\
Dyadic Dialogue & 42 & 113 & 152 & 108 & 33 & 2 & - \\
Multi-party Dialogue & - & 11 & 47 & 27 & 13 & 1 & 1 \\
\bottomrule
\end{tabular}
\caption{Distributions of instructions with different numbers of constraints.}
\label{tab:sub-tasks}
\end{table}

\subsection{Evaluation Protocol}
We employ GPT-4o \cite{DBLP:journals/corr/abs-2303-08774} as the evaluation model to assess constraint adherence in generated responses in LLM-as-judge manner~\cite{zhang2023wider,wang-etal-2024-large-language-models-fair,zeng2024evaluating,wang2024reframing}. Following established practices \cite{DBLP:journals/corr/abs-2407-03978}, the $k$-th constraint in the $j$-th instruction of $i$-th instance is given a binary compliance score \( S_{i,j,k} \in \{0,1\} \), with 1 signifying full compliance and 0 indicating non-compliance.



\textbf{Instruction-Level Accuracy (ILA)} measures the success rate of individual instructions by averaging compliance across all instructions within a single instance. For the $i$-th instance, $m_i$ denotes the number of instructions and $c_{i,j}$ is the number of constraints in the $j$-th instruction. We calculate the average score for $n$ instances as the final metric.
\begin{align}
\text{ILA}_{i} &= \frac{1}{m_i} \sum_{j=1}^{m_i} {\prod_{k=1}^{c_{i,j}} S_{i,j,k}} \\
\text{ILA} &= \frac{1}{n} \sum_{i=1}^{n} \text{ILA}_{i}
\end{align}

\textbf{Constraint-Level Accuracy (CLA)} assesses the fulfillment of individual constraints, making it crucial for identifying specific requirement violations.
\begin{align}
\text{CLA}_{i} & = \frac{1}{\sum_{j=1}^{m_i} c_{i,j}} \sum_{j=1}^{m_i} \sum_{k=1}^{c_{i,j}} S_{i,j,k} \quad   \\
\text{CLA} & = \frac{1}{n} \sum_{i=1}^{n} \text{CLA}_{i} \quad
\end{align}

These metrics progressively assess compliance at different granularities: from strict instruction-level compliance (ILA) to fine-grained constraint-level analysis (CLA).

\subsection{Quality Control}
To ensure high-quality evaluation data, we implement a post-inspection protocol following initial generation. First, we leverage \texttt{Qwen2.5-72B-Instruct} to systematically verify instruction-clarity alignment, logical consistency of constraints, and overall task feasibility, while automatically detecting and correcting identifiable errors through iterative self-refinement. Subsequently, three certified annotation specialists perform manual review to remove redundant constraints and instructions, revise infeasible tasks, and resolve ambiguous phrasing, ensuring both technical rigor and practical usability.

\section{Preliminaries}
Existing Large Reasoning Models (LRMs) \cite{DBLP:journals/corr/abs-2505-02156, jaech2024openai} have demonstrated improved performance on complex tasks via structured reasoning. A pivotal element in training these LRMs is the use of the Group Relative Policy Optimization (GRPO) \cite{shao2024deepseekmath} algorithm in Reinforcement Learning (RL). Given a query $q$, GRPO samples a group of outputs $\{o_1, o_2,\cdots,o_G\}$ from the old policy $\pi_{{\theta}_{old}}$ and eliminates the need for a separate value function and instead relies on the average reward as a baseline to compute the advantage. The optimization process for policy is as follows:
\begin{align}
    \mathcal{J}_{\rm GRPO}&(\theta) = 
    \mathbb{E}_{q \sim P(Q), \{y_i\}_{i=1}^G \sim \pi_{\theta_{\rm old}}(y|q)} \notag\\
    &\Bigg\{  
    \frac{1}{G} \sum_{i=1}^G \frac{1}{|y_i|} \sum_{t=1}^{|y_i|} 
    \Big\{ \min \Big[
    r_{i,t}(\theta) A_{i,t},  \notag \\
    &\text{clip} \big( r_{i,t}(\theta), 1 - \epsilon, 1 + \epsilon \big) A_{i,t} \Big] \notag \\
    & - \beta \mathbb{D}_{\rm KL}\left[\pi_{\theta} || \pi_{\rm ref}\right] \Big\} \Bigg\}, \\
    A_{i,t} = & \frac{r_{i} - \text{mean}(\{r_1,\cdots,r_G\})}{\text{std}(\{r_1,\cdots,r_G\})},
    \label{eq:GRPO}
\end{align}
where $\epsilon$ and $\beta$ are hyper-parameters, $A_{i,t}$ is the advantage for each specific token, $r_{i,t}(\theta)$ represents the probability ratio or importance sampling weight between the new policy $\pi_\theta$ and the old policy $\pi_{\theta_{\rm old}}$ and $\mathbb{D}_{\rm KL}\left[\pi_{\theta} || \pi_{\rm ref}\right]$ denotes the KL divergence between the trained policy and the reference policy. Detailed information is shown in Appendix \ref{grpo}.

\section{Segment Policy Optimization}
\label{section-7}


Although Group Relative Policy Optimization (GRPO) has demonstrated effectiveness in enhancing the performance of Large Reasoning Models (LRMs), existing methods often encounter difficulties in scenarios that require the simultaneous execution of multiple instructions. Specifically, the reliance on group-level advantage computation may constrain the model's effectiveness in fine-grained task settings. To address these challenges, we propose Segment Policy Optimization (\textbf{SegPO}), which incorporates reasoning mechanisms and segment-level evaluation into the advantage computation. This integration enhances instruction alignment, thereby improving both the accuracy and robustness of model outputs.

\textbf{Dual-Component Advantage Estimation}.
In SegPO, the advantage $A_{i,t}$ for the $t$-th token in the response $o_i$ consists of two components: the global advantage $A^o_{i,t}$ and the segment advantage $A^{\phi}_{i,t}$. For the global advantage, we use a group of rewards $\{r^o_1, \cdots,r^o_G\}$ corresponding to the outputs within each group for computation. For the segment advantage $A^{\phi}_{i,t}$, we select the group of rewards $\{r^\phi_{1, I^i_t},\cdots,r^\phi_{G, I^i_t}\}$ for the corresponding $I^i_t$-th instruction in outputs for computation. The process is as follows:
\begin{align}
    A^o_{i,t} = & \frac{r^o_{i} - \text{mean}(\{r^o_1,\cdots,r^o_G\})}{\text{std}(\{r^o_1,\cdots,r^o_G\})}, \\ 
    A^\phi_{i,t} = & \frac{r^\phi_{i, I^i_t} - \text{mean}(\{r^\phi_{1, I^i_t},\cdots,r^\phi_{G, I^i_t}\})}{\text{std}(\{r^\phi_{1, I^i_t},\cdots,r^\phi_{G, I^i_t}\})}, \\
    A_{i,t} = & A^o_{i,t} + A^\phi_{i,t}.
\end{align}

\textbf{Rewards}.
Specifically, we employ both LLM-based and rule-based systems to determine the rewards. For each response \(o_i\) to the query \(q\), \( r^o_i \) captures accuracy and format compliance. Our rule-based system mandates that reasoning is enclosed between `start\_think' and `end\_think' tags, and answers between `start\_answer' and `end\_answer'. The format score, \( r^f_i \), is one if the format is adhered to, otherwise zero. We assess ILA$_{i}$ and CLA$_{i}$ metrics using state-of-the-art LLMs (i.e., \texttt{Qwen2.5-72B-Instruct}), with scores increased if all instructions are correctly executed. Furthermore, for the $t$ token in the response \( o_i \) associated with the $I^i_t$-th instruction, we define the segment reward \( r^\phi_{i,I^i_t} \) as 1 if all the constraints in the $I^i_t$-th instruction are satisfied, else 0. Details of the training template are provided in the Appendix \ref{prompt}. The reward process is summarized as follows:
\begin{align}
    r^o_{i} = \text{ILA}_{i} + \text{CLA}_{i} &+ \prod_{j=1}^{m_i} \prod_{k=1}^{c_{i,j}} S_{i,j,k} + r^f_i, \\
    r^\phi_{i, I^i_t} &= \prod_{k=1}^{c_{i,I^i_t}} S_{i,I^i_t,k}.
\end{align}

\section{Experiments}
\label{section-4}
\subsection{Baselines}

We compare the performance of both proprietary and open-source LLMs trained on diverse corpora. In the proprietary category, we evaluate models such as GPT-4o \cite{DBLP:journals/corr/abs-2303-08774}, GPT-4o-mini \cite{DBLP:journals/corr/abs-2303-08774}, Claude3.5-Sonnet \cite{anthropic2024claude1}, Claude3.5-Haiku \cite{anthropic2024claude}, gemini-1.5-Pro \cite{DBLP:journals/corr/abs-2403-05530}, gemini-2.0-Flash and GPT-o3-mini. Among open-source models, we assess LLaMA3.1 \cite{DBLP:journals/corr/abs-2407-21783}, Qwen2 \cite{DBLP:journals/corr/abs-2407-10671}, Qwen2.5, DeepSeek-V3 \cite{DBLP:journals/corr/abs-2412-19437}, DeepSeek-R1 \cite{DBLP:journals/corr/abs-2403-05530}, QwQ-32B \cite{DBLP:journals/corr/abs-2412-15115}, and Qwen3 \cite{yang2025qwen3} to explore their efficiency. 

\subsection{Settings}
For inference, we efficiently process proprietary models through their APIs. For open-source models, we employ a robust setup consisting of four Nvidia A100 GPUs, each equipped with 80GB of VRAM, utilizing the vLLM framework on \textsc{EifBench} where applicable. This configuration enables the completion of all tasks in roughly 30 minutes. During evaluation, the GPT-4o model serves as the evaluator, with assessment durations ranging from 4 to 10 hours based on task complexity. Our code and dataset are publicly available for reproducibility.\footnote{Available at \url{https://github.com/Hope-Rita/EIFBench} and \url{https://github.com/Tongyi-CCAI/Complex-IF}.}



\subsection{Results Analysis}


\begin{table*}[t]
\renewcommand\arraystretch{1.1}
    \centering
    \begin{tabular}{lcccccc}
        \toprule
        \textbf{Model} & \multicolumn{2}{c}{\textbf{Plain Text}} & \multicolumn{2}{c}{\textbf{Dyadic Dialogue}} & \multicolumn{2}{c}{\textbf{Multi-party Dialogue}} \\
        \cmidrule(lr){2-3} \cmidrule(lr){4-5} \cmidrule(lr){6-7}
        & ILA $\uparrow$ & CLA $\uparrow$ & ILA $\uparrow$ & CLA $\uparrow$ & ILA $\uparrow$ & CLA $\uparrow$ \\
        \midrule
        \emph{Closed-Source LLMs} & & & & & & \\
        GPT-4o & \textbf{0.2480} & 0.6518 & 0.2166 & 0.5631 & 0.2226 & 0.5786 \\
        Claude-3.5-Sonnet & 0.0896 & 0.3951 & 0.0919 & 0.4142 & 0.0663 & 0.3865 \\
        GPT-4o-mini & 0.0826 & 0.5299 & 0.0930 & 0.4892 & 0.0952 & 0.6001 \\
        Claude-3.5-Haiku & 0.0332 & 0.2214 & 0.0251 & 0.1613 & 0.0142 & 0.1081 \\
         gemini-1.5-Pro & 0.1669 & 0.6705 & 0.2717 & 0.7461 & 0.1972 & 0.7693 \\
         gemini-2.0-Flash & 0.2291 & {0.7028} & {0.1813} & 0.5779 & 0.1681 & {0.5383} \\
        GPT-o3-mini & 0.1743 & 0.7210 &0.0805  & 0.5901  & \textbf{0.3326} & \textbf{0.8672} \\
        \midrule
        \emph{Open-Source LLMs} & & & & & & \\
        LLaMA3.1-8B-Instruct & 0.0127 & 0.2918 & 0.0069 & 0.1845 & 0.0024 & 0.2898 \\
        LLaMA3.1-70B-Instruct & 0.0222 & 0.3696 & 0.0250 & 0.3297 & 0.0156 & 0.3774 \\
        Qwen2-7B-Instruct & 0.0261 & 0.3531 & 0.0269 & 0.2954 & 0.0136 & 0.3666 \\
        Qwen2-72B-Instruct & 0.0823 & 0.5924 & 0.1336 & 0.6458 & 0.0878 & 0.6345 \\
        Qwen2.5-7B-Instruct & 0.0503 & 0.5051 & 0.0742 & 0.5526 & 0.0572 & 0.5878 \\
        Qwen2.5-72B-Instruct & 0.1983 & {0.7565} & \underline{0.2787} & {0.7657} & \underline{0.2636} & \underline{0.8308} \\
        QwQ-32B & 0.0884 & 0.4724 & 0.0909 & 0.4220 & 0.0820 & 0.5439 \\
        DeepSeek-V3 & 0.1955 & 0.6836 & 0.1864 & 0.7206 & 0.1664 & 0.7395 \\
        DeepSeek-R1 & \underline{0.2219} & 0.6860 & \textbf{0.3486} & \textbf{0.7906} & 0.2251 & 0.7465 \\
        Qwen3-32B & 0.2050 & \underline{0.7694} & 0.2513 & \underline{0.7799} & 0.2299 & 0.8078 \\
        Qwen3-32B w/o thinking & 0.2073 & \textbf{0.7703} & 0.2396 & 0.7794 & 0.2119 & 0.7445 \\
        Qwen3-235B-A22B & 0.1700 & 0.6712 & 0.2328 & 0.7296 & 0.2120 & 0.7462 \\
        Qwen3-235B-A22B w/o thinking & 0.1775 & 0.6692 & 0.2252 & 0.7282 & 0.2011 & 0.7444 \\
        \bottomrule
    \end{tabular}
    \caption{Performance metrics across different task categories: Plain Text, Dyadic Dialogue, and Multi-party Dialogues. The best and second-best results are highlighted in bold and underlined.}
    \label{tab:updated_combined_performance}
\end{table*}

\begin{table*}[t]
\renewcommand\arraystretch{1.1}
    \centering
    \begin{tabular}{lcccccc}
        \toprule
        \textbf{Model} & \multicolumn{2}{c}{\textbf{Plain Text}} & \multicolumn{2}{c}{\textbf{Dyadic Dialogue}} & \multicolumn{2}{c}{\textbf{Multi-party Dialogue}} \\
        \cmidrule(lr){2-3} \cmidrule(lr){4-5} \cmidrule(lr){6-7}
        & ILA $\uparrow$ & CLA $\uparrow$ & ILA $\uparrow$ & CLA $\uparrow$ & ILA $\uparrow$ & CLA $\uparrow$ \\
        \midrule
        Qwen2.5-7B-Instruct & 0.0503 & 0.5051 & 0.0742 & 0.5526 & 0.0572 & 0.5878 \\
        Qwen2.5-7B-Instruct w/ GRPO & 0.1345 & 0.6237 & 0.1591 & 0.6393  & 0.2183 & 0.7392 \\
        Qwen2.5-7B-Instruct w/ SegPO & \textbf{0.1460} & \textbf{0.6693} & \textbf{0.1797} & \textbf{0.6791} & \textbf{0.2713} &  \textbf{0.7727} \\
        \bottomrule
    \end{tabular}
    \caption{SegPO Performance across different task categories compared to GRPO.}
    \label{tab1:updated_combined_performance}
\end{table*}

\subsubsection{How do existing LLMs perform?}
The \textsc{EifBench} evaluation, detailed in Tables \ref{tab:updated_combined_performance}, challenges language models by simulating real-world scenarios across three datasets: plain text tasks, dialogue tasks, and multi-party dialogue tasks. These datasets reflect diverse practical applications, with plain text focusing on simple information processing, dyadic dialogues examining conversational dynamics, and multi-party dialogues showcasing collaborative discussions.

Our evaluation uses two key metrics: Instruction-Level Accuracy (ILA) and Constraint-Level Accuracy (CLA). Recent studies \cite{DBLP:journals/corr/abs-2408-01122, DBLP:journals/corr/abs-2411-06208, DBLP:conf/acl/LiZQLLWLYMZZLZM24} emphasize CLA, which measures models' effectiveness in meeting individual constraints with high accuracy. Yet, ILA reveals challenges, as models often fail to satisfy all constraints of a single instruction, resulting in a low probability of executing all instructions in an instance. This highlights the need to enhance multi-task capabilities for adhering to comprehensive instructions in the challenging contexts of the \textsc{EifBench} dataset.

Model performance varies notably across categories, revealing task-type dependencies. In closed-source models, GPT-4o excels in ILA with relatively lower CLA. This indicates its capability to focus and complete individual sub-tasks effectively, albeit less so on fulfilling comprehensive constraints. Well-performing models in the open-weight landscape generally fall into two categories. The first is generalist models, like Qwen2.5-72B-Instruct, where reasoning emerges from large-scale and instruction tuning. Specialized models like DeepSeek-R1 and Qwen3-32B, equipped with explicit reasoning strategies, demonstrate their advanced capabilities with a distinct performance profile. They achieve substantially superior performance on ILA metrics, while also showing robust, comparable performance on CLA metrics against their non-reasoning counterparts. This highlights that their specialized architecture enables a deeper understanding of complex tasks, leading to success on targeted metrics.

\subsubsection{Effectiveness of SegPO}
We implement the Group Relative Policy Optimization (GRPO) \cite{shao2024deepseekmath} framework, employing the overall reward \( r_o \) as the advantage value to enhance model capabilities. 
As illustrated in \tabref{tab1:updated_combined_performance}, SegPO achieves significant improvements compared to base model and GRPO with respective 14.85\% and 3.40\% average increases, which confirms the effectiveness and necessity of segment-level advantage computation for accurate understanding of multiple task.
The reason is that respectively calculating advantages for the response corresponding to each instruction in the input may result in a more precise reward, effectively steering the model's learning. 

\subsubsection{Full Constraint Satisfaction Analysis}
In real-world scenarios, fully satisfying \textit{all} constraints across \textit{all} instructions is crucial especially for LLM agents with long-horizon decision-making involving multiple tasks, aside from ILA and CLA metrics. Our analysis revealed that the leading performance in dyadic dialogue was achieved by gemini-1.5-Pro and DeepSeek-R1, both scoring 0.0044, with GPT-4o following as the second-best at 0.0022. All other models recorded a performance score of zero. This relatively low performance highlights the increased difficulty posed by our benchmark, which, unlike previous datasets with limited constraints, is crafted to simulate realistic tasks such as smart home operations. These scenarios require handling multiple interdependent constraints simultaneously. The results indicate the current models' limitations in reasoning and executing complex, constraint-rich instructions, emphasizing the need for further advancements in their capabilities.

\begin{table}[t]
    \centering
        \resizebox{\columnwidth}{!}{
\begin{tabular}{lccc}
\toprule
Type & {Human 1} & Human 2 & Human 3 \\ 
\midrule
Plain Text        & 0.9234      & 0.9342      & 0.9083      \\
Dyadic Dialogue   & 0.9341      & 0.9268      & 0.9326      \\
Multi-Party Dialogue & 0.9118      & 0.9021      & 0.9164      \\
\hline
Average           & 0.9231      & 0.9210      & 0.9191      \\
\bottomrule
\end{tabular}}
\caption{PCC Between \texttt{Qwen2.5-72B-Instruct} and expert evaluations on quality assessment.}
\label{tab: generation qa}
\end{table}

\begin{table}[t]
    \centering
        \resizebox{\columnwidth}{!}{
\begin{tabular}{lccc}
\toprule
Type          & Human 1 & Human 2 & Human 3 \\
\midrule
Plain Text        & 0.7123  & 0.7236  & 0.7172  \\
Dyadic Dialogue   & 0.7438  & 0.7632  & 0.7524  \\
Multi-Party Dialogue & 0.7551 & 0.7459  & 0.7376  \\
\hline
Average           & 0.7371  & 0.7442  & 0.7357  \\
\bottomrule
\end{tabular}}
\caption{The kappa coefficient between expert evaluations and GPT-4o-as-Judge in the evaluation process.}
\label{tab: evaluation qa}
\end{table}

\subsection{Quality Assessment}
We validated the benchmark's quality through both data generation and evaluation processes. First, we assessed the dataset from \texttt{Qwen2.5-72B-Instruct} by randomly selecting 50 instances, comparing model scores with evaluations from three experts for contradictions, redundancy, and infeasibility within instructions and constraints. The Pearson Correlation Coefficient (PCC) in \tabref{tab: generation qa} shows strong consistency, supporting benchmark credibility. Additionally, we validated LLM-judge evaluations by comparing them with human assessments across three datasets. We randomly selected 500 constraints per dataset based on LLM-generated responses and calculated Fleiss' Kappa scores \cite{fleiss1971measuring} between the results from GPT-4o-as-judge and human evaluators. High consistency in \tabref{tab: evaluation qa} confirms the reliability of our evaluation process.

\section{Related work}
\label{section-5}

\subsection{Instruction Following} 
Recent advancements in fine-tuning large language models (LLMs) show that annotated instructional data significantly enhances models' ability to comprehend and execute diverse language instructions \cite{DBLP:conf/emnlp/WellerLGP20, DBLP:conf/acl/Ye020, DBLP:conf/acl/MishraKBH22}. Building on this, incorporating more detailed and sophisticated instructions has been shown to further improve model capabilities \cite{lou2023comprehensive}. For instance, \cite{DBLP:conf/iclr/XuSZG0FTLJ24} presents a method of incrementally generating complex instructions from seed instructions using LLMs, enabling LLaMA to surpass 90\% of ChatGPT’s performance in 17 out of 29 skills. Additionally, research is increasingly focusing on constrained instructions \cite{DBLP:journals/corr/abs-2404-02823, DBLP:journals/corr/abs-2406-13542, DBLP:conf/emnlp/HeZHLX24}, a subset of complex instructions, aimed at enhancing models' ability to handle intricate challenges by increasing instructional constraints.




\subsection{Evaluation of Instruction Following} 
Instruction following significantly impacts the effectiveness of large language models (LLMs) \cite{DBLP:journals/corr/abs-2308-05374}. Early work focused on evaluating compliance with simple directives, often involving single constraints like semantic \cite{DBLP:conf/nips/ZhengC00WZL0LXZ23, DBLP:conf/acl/LiuLWHFWCKXTZ0G24} or formatting \cite{DBLP:conf/acl/XiaXDYFX0X24, DBLP:conf/naacl/TangZPZZCG24} requirements. As LLMs find their way into more complex real-world applications, the need to assess their capacity to handle sophisticated instructions has grown \cite{DBLP:conf/acl/QinSHYCWW00Y24, DBLP:conf/acl/Jiang0ZZLMS00W24}. For example, \citep{DBLP:journals/corr/abs-2404-02823} introduced the Conifer dataset to enhance LLMs' handling of multi-level instructions with complex constraints, while \citep{DBLP:conf/acl/QinSHYCWW00Y24} designed a method for decomposing single instructions into multiple constraints. Moreover, \citep{DBLP:conf/aaai/HeZHCXHZLX24} created benchmarks using real-world constraints, and \citep{DBLP:journals/corr/abs-2407-03978} further innovated by integrating diverse constraint types. 
Despite these advancements, current datasets often lack the extensive constraints seen in multi-instruction, multi-constraint real-world scenarios.

\section{Conclusion}

In conclusion, this study introduces the \textbf{E}xtremely Complex \textbf{I}nstruction \textbf{F}ollowing \textbf{Bench}mark (\textbf{\textsc{EifBench}}), addressing existing single-task dataset limitations by incorporating multi-task scenarios and constraints for realistic evaluation of large language models (LLMs). We also propose the \textbf{Seg}ment \textbf{P}olicy \textbf{O}ptimization (\textbf{SegPO}) algorithm algorithm, which enhances LLMs' multi-task workflow execution, showing a 14.85\% improvement on \textsc{EifBench} over \texttt{Qwen2.5-7B-Instruct}. Evaluations reveal significant performance gaps, highlighting the need for models capable of tackling real-world complexities. This benchmark sets a new standard, steering future research toward developing robust and adaptable systems for practical applications.
\section*{Limitations}
While \textsc{EifBench} provides a robust evaluation framework for plain text, dyadic dialogue, and multi-party tasks, it has two limitations that could be addressed in future work. First, the inter-task relationships could be further enhanced to reflect more complex, real-world dependencies, such as multi-step reasoning or conditional task execution. Second, the dataset currently focuses primarily on Chinese instructions, which limits its applicability to multilingual scenarios. Expanding to include more languages would improve its global relevance and enable evaluation of LLMs' cross-lingual capabilities. Addressing these limitations would make \textsc{EifBench} even more comprehensive and aligned with practical applications.

\section*{Acknowledgments}
We would like to thank the reviewers for their helpful reviews and feedback. This work was supported by Alibaba Research Intern
Program.
\bibliography{custom}
\clearpage
\appendix
\section{Taxonomy of Constraint}
We present the taxonomy of constraint in \tabref{tab:constraints}.
\label{app:con}
\begin{table*}[t]
    \centering

    \begin{tabular}{ll}
        \toprule
        \textbf{Constraint Type} & \textbf{Constraint Dimension} \\
        \midrule
        Content Constraint & Theme Constraint \\
                           & Exclusion Constraint \\
                           & Inclusion Constraint \\
                           & Value Constraint \\
                           & Privacy Constraint \\
                           & Numerical Constraint \\
        \cmidrule(lr){1-2}
        Situation Constraint & Role-Playing Constraint \\
                             & Target Audience Constraint \\
                             & Prior Condition Constraint \\
                             & Natural Language Process Background Information Constraint \\
                             & Markdown Process Background Information Constraint \\
                             & Table Background Information Constraint \\
                             & Text Background Information Constraint \\
        \cmidrule(lr){1-2}
        Style Constraint & Tone and Style Constraint \\
                         & Emotion Constraint \\
                         & Linguistic Characteristics Constraint \\
                         & Multilingual Constraint \\
        \cmidrule(lr){1-2}
        Format Constraint & Output Format Constraint \\
                          & Text Pattern Constraint \\
                          & Grammar Structure Constraint \\
                          & Citation Constraint \\
                          & Numbering and List Constraint \\
                          & Hierarchical Structure Constraint \\
                          & Template Constraint \\
        \bottomrule
    \end{tabular}
    \caption{Constraints and Their Dimensions}
    \label{tab:constraints}
\end{table*}

\section{Detailed Information on GRPO}
\label{grpo}
The ratio $r_{i,t}(\theta)$ represents the probability ratio or importance sampling weight between the new policy $\pi_\theta$ and the old policy $\pi_{\theta_{\rm old}}$:
\begin{align}
    r_{i,t}(\theta) = \frac{\pi_\theta(o_{i,t} | q, o_{i,<t})}{\pi_{\theta_{\rm old}}(o_{i,t} | q, o_{i,<t})},
    \label{eq:rit-obj}
\end{align}
and GRPO estimates the KL divergence with the following unbiased estimator:
\begin{align}
    \mathbb{D}_{\rm KL}\left[\pi_{\theta} || \pi_{\rm ref}\right] &= 
    \frac{\pi_{\rm ref}(o_{i,t}|q,o_{i,<t})}{\pi_{\theta}(o_{i,t}|q,o_{i,<t})} \notag \\
    &- \log\frac{\pi_{\rm ref}(o_{i,t}|q,o_{i,<t})}{\pi_{\theta}(o_{i,t}|q,o_{i,<t})} - 1.
\end{align}

\section{Experiment Analysis}
\subsection{Factors Influencing Instruction Following}
\begin{figure*}[!tbp]
    \centering
    \includegraphics[width=1.8\columnwidth]{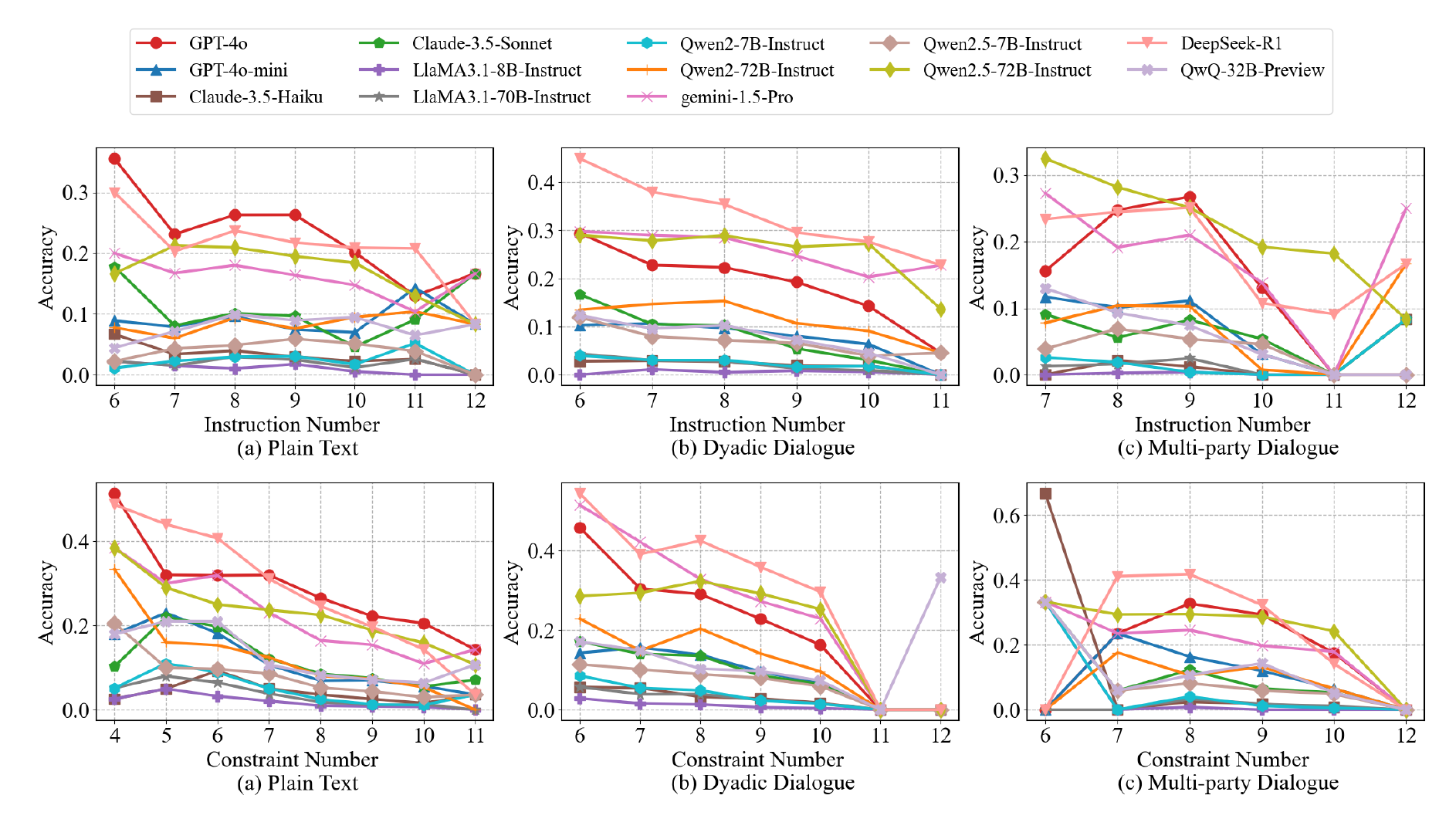}
    \caption{Performance on different numbers of instructions and constraints.}
    \label{fig:analysis}
\end{figure*}
To conduct our investigation, we sampled both open-source and closed-source models with varying performance levels—some exemplary and others average—and visualized their results. Our investigation identifies two critical dimensions influencing instruction adherence in language models: (1) the number of instructions per instance and (2) the number of constraints per instruction. As illustrated in \figref{fig:analysis}, performance degrades progressively as these variables increase, though with minor patterns. This decline is particularly pronounced with an increase in constraints, likely because each additional constraint raises the complexity of completing a task, making it more challenging for the model to meet all requirements. Conversely, the interdependence between instructions is generally low, meaning that an increase in the number of instructions does not lead to as steep a performance decline. This is primarily because the difficulty lies in managing multiple tasks simultaneously, rather than the instructions themselves being interrelated. In some instances, especially where there are larger numbers of instructions and constraints, performance may inexplicably improve. This can be attributed to the smaller sample sizes in these scenarios, leading to greater variability in performance outcomes. Overall, this analysis underscores the intricacies of maintaining consistent instruction adherence across diverse scenarios.

\section{Data Instance}
In this section, we present an example instance to illustrate the application and analysis of the idiom.

\begin{quote}
\tt\small
Instruction\_0: "Explain the origin and significance of the Chinese idiom 'drawing legs on a snake' (huà shé tiān zú)"

Constraints:
\begin{itemize}
\item Must provide a detailed account of the idiom's historical background and origin.
\item Avoid using the words "meaning" or "explanation" to describe its significance.
\item Follow the context → story → implications structure.
\end{itemize}

Instruction\_1: "Create a sentence containing the idiom 'drawing legs on a snake' "

Constraints:
\begin{itemize}
\item The sentence must be 20-30 Chinese characters long.
\item The sentence must be non-declarative (e.g., rhetorical question, exclamation, or imperative).
\end{itemize}

Instruction\_2: "Analyze the specific scenario of 'drawing legs on a snake' in your created sentence."

Constraints:
\begin{itemize}
\item Describe in detail the superfluous action within the scenario.
\item Include a root-cause analysis of why this "unnecessary addition" leads to negative consequences.
\end{itemize}
\end{quote}

\clearpage
\twocolumn

\section{Prompt}
\label{prompt}
\subsection{Prompt for task expansion}

\begin{quote}
\tt\small
You are an assistant to help generate comprehensive multi-task tasks from basic tasks/basic texts/basic dialogues. Based on the given basic task, please design 5-10 different types of extended tasks, which must be reasonable and meet actual needs. The generated tasks should be placed after ``-output-:''.\\

Please follow these rules when generating tasks:\\

1. Task design must be based on the input text content or the already designed task output content.\\

2. Task instructions should be clear and specific.\\

3. Each task should include explicit output format requirements.\\

5. Aim to increase task difficulty, selecting tasks that require multi-step reasoning and thinking.\\

6. Tasks should be related; specific task details can vary. The connections can be selective, sequential, parallel, etc. At least three types of connections are needed, including:\\
   A. Parallel Task Mode: Analyzing multiple dimensions simultaneously \\
   B. Sequential Task Mode: Task chain dependency\\
   C. Conditional Selection Mode: Branch based on different situations, considering possible branches of the task, and design different tasks for different branches \\
   D. Nested Task Mode: Hierarchical task structure\\
7. Task Design Principles:\\
   - Clear goals\\
   - Clear instructions\\
   - Specific steps\\
   - Standardized format\\
   - Evaluability\\
8. Task types may repeat, but task content may not.\\

9. Note that expansion must be based on the given basic task, expanding into richer, more comprehensive, and varied integrated tasks. The text material in the given basic task must be retained, as subsequent tasks will all involve it!\\

10. Write the extended tasks after ``-output-:'', and the thought process and analysis for generating the extended tasks after ``-explanation-:''.\\

--input--:\\
\{text\}\\

-output-:\\

-explanation-:\\
...\\

Task-type examples should be chosen from the following categories. The specific task examples are listed in each task category. Note that you are only required to design tasks, not provide example outputs:\\

1. Classification Task \\
- Sentiment Analysis \\
- Text Classification \\
- Toxic Content Detection \\
- Empathy Detection \\
- Stereotype Detection\\
- Social Norm Judgment\\

2. Information Extraction\\
- Named Entity Recognition \\
- Keyphrase Annotation \\
- Coreference Resolution \\
- Entity Relationship Classification\\

3. Text Generation
- Story Creation\\
- Poetry Generation\\
- Recipe Generation\\
- Outline Generation\\
- Text Expansion/Compression\\
- Title Generation\\
- Data Description Generation\\
- Text Rewriting/Simplification\\

4. Dialogue Systems\\
- Dialogue Generation\\
- Intent Recognition\\
- Question Generation/Rewriting\\
- Dialogue State Tracking\\
- Role-playing Dialogue\\

5. Reasoning and Logic\\
- Common Sense QA\\
- Multi-hop QA\\
- Critical Thinking Judgment\\
- Mathematical Reasoning\\
- Theory of Mind Reasoning\\

6. Language Style\\
- Style Transfer\\
- Language Detection\\
- Sarcasm Detection\\
- Spelling/Punctuation Error Detection\\

7. Evaluation and Verification\\
- Text Quality Evaluation\\
- Fact-checking\\
- Answer Verification\\
- Uncertainty Judgment\\

8. Programming-related\\
- Code Generation/Debugging\\
- Code Explanation\\
- Code Translation\\

Each example format is as follows:\\
Task x:\\
Type:\\
Specific Requirements:\\
- ...\\
- ...\\
\end{quote}

\subsection{Prompt for task revision}
\begin{quote}
\tt\small
You are a task optimization expert. Please analyze and optimize the given task set.

Input text and tasks are as follows:
\begin{verbatim}
--input--:
query: {input_text}
task: {task}
\end{verbatim}

First, output all optimized tasks (if there are no modifications, output the original tasks) in Chinese after ``-output-''. Secondly, write the optimized rationale and analysis process after``-explanation-''. Please strictly follow this format.

The output format is as follows:

\begin{verbatim}
-output-:
Task 1: ...
Task 2: ...

-explanation-:
\end{verbatim}

Please follow these steps for analysis and optimization: \\
1. Input Analysis \\
Input Type Judgment: \\
- Determine if it's a complete text or a task requirement \\
- Check if it includes a creative/analytical directive \\
- Assess the amount of information provided by the text/task \\
- Preserve textual information if input text analysis is involved \\

2. Task Reasonability Check
Analyze each task: \\
Reasonability of Task X: \\
A. Matching Degree with Input\\
- Does the task rely on the actual provided information, and is there excessive speculation or extension?\\

B. Executability of the Task\\
- Is there sufficient information to support it, and are the scoring criteria operable?\\

C. Existing Problems\\
- [List specific problems]\\

3. Modification Suggestions\\
If modification is required, provide specific modification directions and design of revised tasks, and modify the tasks according to this suggestion.
\end{quote}

\subsection{Prompt for task combination}
\begin{quote}
\tt\small
You are an integration assistant for input and task requirements. Your goal is to combine the basic tasks (including tasks and reading materials) given in ``--input--'': and the expanded tasks in ``--task--'': to generate comprehensive tasks that include reading material information and task information.
Please note that the integrated tasks will not fetch text information from elsewhere, so ensure that the generated comprehensive tasks include the text material.
Please integrate all tasks from the expanded tasks into the comprehensive tasks. Identify all expanded tasks and ensure the number of sub\_instructions matches the number of sub-tasks in the expanded tasks.
Please ensure to extract and integrate the materials and text information involved in --input--:, and do not omit any details.

\begin{verbatim}
--input--:
{input_text}
--tasks--:
{task}
\end{verbatim}

Please put the generated content in Chinese after ``-output-:'', including:
1. First, place the comprehensive task and the text materials involved in the task after ``INSTRUCTION:''. Please ensure to fully include the reading materials from the basic tasks.
2. Then, sequentially output all sub-instructions, each starting with ``SUB\_INSTRUCTION\_X:'', including ``instruction:'' and ``constraints:'' parts. After ``instruction:'', write the content of the sub-instruction, and after ``constraints:'', write several constraint items. Each constraint should follow the format ``- constraint content [constraint type]''.
Lastly, provide the specific combination process after ``-explanation-:''.

The output format is as follows:
\begin{verbatim}
-output-:
INSTRUCTION:
...
SUB_INSTRUCTION_x:
instruction: ...
constraints:
- ... [...]
- ... [...]

--explaination--:
...
\end{verbatim}

Please follow these steps for analysis and combination:

1. Input Analysis\\
- Extract **original tasks** and corresponding **text materials** from --input-- \\
- Extract specific text content and basic requirements \\
- Extract all specific requirements from the input text \\
- If the input involves text, it must be placed in the comprehensive task to avoid missing the input text

2. Task Expansion Analysis \\
- Extract **sub-task** information (information type, information volume, target) from --task-- \\
- Retrieve related expanded tasks (such as information extraction, reasoning, etc.) \\
- Understand the relevance and progression relationship between tasks \\
- Identify all constraints and restrictions \\
- Record keywords and special conditions \\
- Note that all tasks are prefixed with ``Task'', be sure to identify all tasks, and the number of tasks should match the number of sub\_instructions \\
- List all tasks and their sub-tasks

3. Combine into a New Comprehensive Task \\
- Expand **original tasks**, **text materials**, and all **sub-tasks** into a comprehensive analysis task, and output to INSTRUCTION \\
- Maintain logical connections between tasks \\
- Ensure the INSTRUCTION meets all sub-task requirements \\
- Make sure to integrate all tasks and incorporate the input \\
- Ensure all original requirements are covered

4. Integrated Output \\
- A unified main instruction, output the new comprehensive task (INSTRUCTION): \\
  * Use natural language connectors, the task requirements should be connected with natural language, such as ``then'', ``next'', ``finally'', to maintain fluency \\
  * Ensure the integration of input and task, the combined content should be output to INSTRUCTION, forming a coherent and smooth instructional language \\
  * Ensure all requirements are covered

- A series of sub-instructions (SUB\_INSTRUCTION) \\
  * Each sub-instruction contains specific tasks (instruction) \\
  * Each sub-instruction includes specific constraints (constraints), generating 5-10 specific constraints \\
  * The purpose of the constraints is to complete the task as much as possible, the more detailed, the better, with difficulty ranging from simple to complex \\
  * Note that constraints should not be ambiguous or unclear \\
  * Each constraint and its type should be selected from the following 24 types: \\

\begin{itemize}
  \item Theme Constraint
  \item Exclusion Constraint
  \item Inclusion Constraint
  \item Value Constraint
  \item Privacy Constraint
  \item Numerical Constraint
  \item Role-Playing Constraint
  \item Target Audience Constraint
  \item Prior Condition Constraint 
  \item Natural Language Process Background
  \item Markdown Process Background
  \item Table Background Information
  \item Text Background Information
  \item Tone and Style Constraint
  \item Emotion Constraint
  \item Linguistic Characteristics
  \item Multilingual Constraint
  \item Output Format Constraint
  \item Text Pattern Constraint
  \item Grammar Structure Constraint
  \item Citation Constraint
  \item Numbering and List Constraint
  \item Hierarchical Structure Constraint
  \item Template Constraint
\end{itemize}

Precautions: \\
1. Sub-tasks must be clearly mentioned in the integrated instruction. \\
2. Do not change the wording and expressions of the original instructions. \\
3. Split according to the order in which the tasks appear in the instructions. \\
4. Each sub-task is equipped with 5-10 constraint items, with constraint types selected from the above 24 types. \\
5. When integrating, ensure that new tasks are organically combined with the original content, i.e., do not generate instructions based only on information from input.
\end{quote}

\subsection{Prompt for constraint expansion}
\begin{quote}
\tt\small
You are an expert at generating constraints.\\
Please modify the original constraint information for each instruction.\\
For every SUB\_INSTRUCTION, generate 6-10 high-quality constraints.\\
Each constraint must address key requirements of the task with measurable analysis rather than general statements.\\

Input information is as follows:
\begin{verbatim}
--input--:
{input_text}
\end{verbatim}

When outputting content, please place the generated content after ``-output-'' first.\\
Begin with ``INSTRUCTION'', followed by each sub-instruction sequentially, with each sub-instruction starting with ``SUB\_INSTRUCTION\_X'', including both ``instruction'' and ``constraints''.\\
After ``instruction'', write the content of the sub-instruction, and after ``constraints'', write several constraint items.\\
Each constraint should follow the format ``- constraint content [constraint type]''.\\

Finally, place the analysis of the modification process after -explanation-.\\

The generated format is as follows:
\begin{verbatim}
-output-:
INSTRUCTION:
...

SUB_INSTRUCTION_0: 
instruction: ...
constraints: 
- ... [...]
- ... [...]
...

SUB_INSTRUCTION_1: 
instruction: ...
constraints: 
- ... [...]
...

--explaination--:
...
\end{verbatim}

Specific modification requirements:\\
1. Each constraint must be specific and clear, avoiding vague expressions, and the constraint structure should use ``and,'' ``or,'' ``not'' types.\\
2. Each constraint must include measurable standards, such as specific numbers, clear criteria, etc.\\
Also, note that these constraints are for the model to follow, avoiding situations that are impossible to assess, such as ``Please respond within 5 seconds after reading,'' which cannot be evaluated for compliance.\\
3. Avoid generic vocabulary; examples below:\\
Avoid using generic words often found in constraints:\\

Quality Descriptors:\\
``appropriate,'' ``suitable,'' ``adequate,'' ``sufficient,'' ``complete,'' ``detailed,'' ``accurate,'' ``clear,'' ``varied''

Logical Descriptors:\\
``logicality,'' ``coherent,'' ``orderly,'' ``hierarchical,'' ``structured,'' ``systematic'' 

Effect Descriptors:\\
``comprehensive,'' ``practical,'' ``vivid,'' ``specific,'' ``pictorial,'' ``persuasive,'' ``effective,'' ``helpful''

Standard Descriptors:\\
``meets requirements,'' ``standard-compliant,'' ``sufficient,'' ``as stipulated,'' ``qualified,'' ``standard fit''

Feature Descriptors:\\
``characteristic,'' ``feature,'' ``prominent,'' ``obvious,'' ``outstanding''

These words should be replaced with specifically measurable standards, for example:\\
``suitable'' -> ``must include 3 specific examples''\\
``detailed'' -> ``no less than 100 words''\\
``vivid'' -> ``must use more than 3 figures of speech''\\
``logicality'' -> ``must follow [cause-process-result] order''\\
``persuasive'' -> ``must cite 1 authoritative data source''\\

4. Each constraint must be of one of the following types:
\begin{itemize}
  \item Theme Constraint
  \item Exclusion Constraint
  \item Inclusion Constraint
  \item Value Constraint
  \item Privacy Constraint
  \item Numerical Constraint
  \item Role-Playing Constraint
  \item Target Audience Constraint
  \item Prior Condition Constraint
  \item Natural Language Process Background
  \item Markdown Process Background
  \item Table Background Information
  \item Text Background Information
  \item Tone and Style Constraint
  \item Emotion Constraint
  \item Linguistic Characteristics
  \item Multilingual Constraint
  \item Output Format Constraint
  \item Text Pattern Constraint
  \item Grammar Structure Constraint
  \item Citation Constraint
  \item Numbering and List Constraint
  \item Hierarchical Structure Constraint
  \item Template Constraint
\end{itemize}
\end{quote}

\subsection{Prompt for constraint revision}
\begin{quote}
\tt\small
You are an assistant for modifying constraints.\\
Please analyze the original constraint information in the instruction for potential issues, and modify the constraints for each SUB\_INSTRUCTION to generate 6-10 high-quality constraints.\\
Each constraint must address specific, measurable requirements for the task, rather than general statements.\\

\begin{verbatim}
--input--:
{input_text}
\end{verbatim}

When outputting content, first combine the modification analysis process and output the modified content (if no modifications, output the original content) after ``-output-'', including INSTRUCTION and the modified SUB\_INSTRUCTION information, where each SUB\_INSTRUCTION consists of instruction and constraints.\\
Each constraint should follow the format ``- specific constraint content [constraint type]''.\\
Finally, provide the analysis of the modification process after -explanation-.

The generated format is as follows:
\begin{verbatim}
-output-:
INSTRUCTION:
...

SUB_INSTRUCTION_x: 
instruction: ...
constraints: 
- ... [...]
- ... [...]
...

-explanation-:
...
\end{verbatim}

Please follow these steps for analysis and modification:\\

1. Examine each instruction for the following 8 types of issues and modify any issues found

1.1 Vague constraints/lack of specific evaluation metrics need to be detailed into evaluable metrics\\
Example:\\
Original constraint:\\
- The article structure must be reasonable

Modified to:\\
- The article must include introduction, analysis, and conclusion sections, with each section not less than 200 words

Below are frequently used vague words that should be avoided:\\
Quality Descriptors: ``appropriate'', ``suitable'', ``adequate'', ``sufficient'', ``complete'', ``detailed'', ``accurate'', ``clear'', ``varied''\\
Logical Descriptors: ``logicality'', ``coherent'', ``orderly'', ``hierarchical'', ``structured'', ``systematic''\\
Effect Descriptors: ``comprehensive'', ``practical'', ``vivid'', ``specific'', ``pictorial'', ``persuasive'', ``effective'', ``helpful''\\
Standard Descriptors: ``meets requirements'', ``standard-compliant'', ``sufficient'', ``as stipulated'', ``qualified'', ``standard fit''\\
Feature Descriptors: ``characteristic'', ``feature'', ``prominent'', ``obvious'', ``outstanding''

First, check if any vague words appear in the constraints, then refine the vague constraints into evaluable metrics based on the specific task context. Here are some examples:\\
``suitable'' -> ``must include 3 specific examples''\\
``detailed'' -> ``no less than 100 words''\\
``vivid'' -> ``must use more than 3 figures of speech''\\
``logicality'' -> ``must follow [cause-process-result] order''\\
``persuasive'' -> ``must cite 1 authoritative data source''\\
``rich emotional color'' -> Use at least two rhetorical devices (parallelism, contrast, metaphor, personification, or exaggeration) to express emotions\\

1.2 Duplicate constraints need to be distinguished\\
Example:\\
Original constraint:\\
- Must use formal language\\
- Must use standard language\\
Problem: The two constraints are similar and lack distinction\\
Modification suggestion:\\
- Must use honorific words like ``you, your, respectfully''\\
- Must avoid using interjectory words like ``oh, ah, um''\\

1.3 Logical contradictions\\
Example:\\
Original constraint:\\
- Both ``relaxed and gentle'' and ``professional terminology'' require a remedy\\
Suggested modification:\\
- The tone must be friendly and professional, with easy-to-understand explanations provided for professional terminology\\

1.4 Lack of key constraints\\
Example:\\
E-commerce customer service scenario\\
Suggested modification:\\
- Must explain the shop's specific compensation plan\\
- Must provide direct contact details for customer service\\
- Must specify the follow-up timeline\\

Original constraint:\\
- Modify according to the following format\\
Modification suggestion:\\
- Modify according to the table format\\

1.5 Contradictory constraints:\\
Original constraint:\\
- Requires classical Chinese style\\
- Requires vividness\\
Suggestion: Adjust to:\\
- Use classical vocabulary but ensure modern readers can understand\\
- Provide modern explanations for each term\\

1.6 Lack of key definitions:\\
Original constraint:\\
- The calculation of the number of ``events'' lacks a clear definition\\
Suggestion: Add:\\
- Clearly define ``event'' as ``an independent action and its corresponding object''\\
- Provide specific examples for event judgment\\

1.7 Data source missing\\
Original constraint:\\
- ``Must include specific data or factual references''\\
- ``Must be based on specific data and facts''\\
But no instructions on how to obtain and verify data sources\\
Suggestion: Add data source requirements:\\
- ``Must cite authoritative market research agencies or official publications, and specify the source''\\
- ``Data must be from statistical results within the past 2 years''\\

1.8 Evaluation criteria unclear:\\
Original constraint:\\
- ``Applicable scenarios must be reasonable and consistent with market reality''\\
- ``Usage suggestions must be specific and feasible''\\
But no evaluation criteria provided\\
Suggestion: Set specific evaluation indicators:\\
- ``Each suggestion must include usage scenarios, expected effects, and cost considerations''\\
- Add feasibility verification:\\
- ``Each suggestion must be supported by actual cases''\\

2. Constraint types should be selected from the following 24 categories while varying the types as much as possible to enrich the diversity of the constraints:
\begin{itemize}
  \item Theme Constraint
  \item Exclusion Constraint
  \item Inclusion Constraint
  \item Value Constraint
  \item Privacy Constraint
  \item Numerical Constraint
  \item Role-Playing Constraint
  \item Target Audience Constraint
  \item Prior Condition Constraint
  \item Natural Language Process Background
  \item Markdown Process Background
  \item Table Background Information
  \item Text Background Information
  \item Tone and Style Constraint
  \item Emotion Constraint
  \item Linguistic Characteristics
  \item Multilingual Constraint
  \item Output Format Constraint
  \item Text Pattern Constraint
  \item Grammar Structure Constraint
  \item Citation Constraint
  \item Numbering and List Constraint
  \item Hierarchical Structure Constraint
  \item Template Constraint
\end{itemize}
\end{quote}

\subsection{Prompt for constraint combination}
\begin{quote}
\tt\small
Now you are an assistant in integrating tasks and constraints; please help me optimize this comprehensive task's instruction (INSTRUCTION), sub-instructions (SUB\_INSTRUCTION), and constraints.\\
Please ensure that the input information/reading materials in the INSTRUCTION are retained; otherwise, subsequent tasks cannot be completed.\\

Input information is as follows:
\begin{verbatim}
--input--:
{input_text}
\end{verbatim}

First, output the modified comprehensive task content (if no modifications, output the original content) after ``-output-'', including INSTRUCTION and the modified SUB\_INSTRUCTION information. Each SUB\_INSTRUCTION consists of instruction and constraints, with each constraint structured as follows: ``- specific content [constraint type]''.\\
Finally, put the specific modification analysis process after -explanation-.

The output format is as follows:
\begin{verbatim}
-output-:
INSTRUCTION:
...

SUB_INSTRUCTION_0:
instruction: ...
constraints:
- ... [...]
- ... [...]
...

SUB_INSTRUCTION_1:
instruction: ...
constraints:
- ... [...]
- ... [...]
...

-explanation-:
...
\end{verbatim}

Please follow these steps for analysis and modification:\\

1. Analyze the existing instructions and constraints for issues: \\
- Check if the structure is reasonable \\
- Identify duplicate or contradictory requirements \\
- Discover vague or non-executable constraints\\
- Find missing key requirements

2. Provide update suggestions:\\
- Instruction update: Make it clearer and more targeted for the comprehensive task\\
- Sub-instruction update: Specify each atomic task\\
- Constraint update: Provide executable and verifiable constraints\\
- Start with -output-, output the modified instructions (INSTRUCTION), sub-instructions (SUB\_INSTRUCTION), and constraints \\
- Each constraint includes two parts: content [type]

3. When modifying, be sure to keep the input information such as reading materials in the original INSTRUCTION. Do not delete specific query information, causing text errors.

4. Each constraint type must be one of the following, if it is not among these types, please modify the constraint type to one of the following types. If it cannot be modified, regenerate constraints that meet these types:
\begin{itemize}
  \item Theme Constraint
  \item Exclusion Constraint
  \item Inclusion Constraint
  \item Value Constraint
  \item Privacy Constraint
  \item Numerical Constraint
  \item Role-Playing Constraint
  \item Target Audience Constraint
  \item Prior Condition Constraint 
  \item Natural Language Process Background
  \item Markdown Process Background
  \item Table Background Information
  \item Text Background Information
  \item Tone and Style Constraint
  \item Emotion Constraint
  \item Linguistic Characteristics
  \item Multilingual Constraint
  \item Output Format Constraint
  \item Text Pattern Constraint
  \item Grammar Structure Constraint
  \item Citation Constraint
  \item Numbering and List Constraint
  \item Hierarchical Structure Constraint
  \item Template Constraint
\end{itemize}
\end{quote}

\subsection{Prompt for instruction-level validation}
\begin{quote}
\tt\small
You are now an assistant to modify sub-tasks.\\
You need to modify the given sub-tasks according to the following steps:\\

1. Analyze the relationship between sub-tasks and evaluate their role in the comprehensive task, removing contradictory sub-tasks.\\
2. Note that sub-tasks are carried out by a large model, so remove tasks that the large model cannot complete, such as internet searches, finding related information, statistical data analysis, etc.\\
3. Delete tasks with weak logical connections. The relationships between sub-tasks can be:
  A. Parallel Task Mode: Analyzing multiple dimensions simultaneously\\
  B. Serial Task Mode: Chain-dependent tasks\\
  C. Conditional Selection Mode: Branching based on different situations, considering possible branches of a task, and designing different tasks for different branches\\
  D. Nested Task Mode: Hierarchical task structure\\
4. Sub-task selection criteria:
- Remove tasks that an AI model cannot accomplish (such as network searches, finding information)\\
- Remove tasks with weak logical connections\\
- Remove redundant, contradictory, or unreasonable tasks\\
- Optimize sub-task content to be of moderate difficulty and meet practical needs\\
- It is acceptable to propose some creative tasks\\
- Ensure that the number of generated sub-tasks is between 6 and 14\\
- Try to diversify task types, with at least 3 different styles of tasks\\
- Remove tasks that require an AI model to use tools, such as Named Entity tools, etc.\\

5. Select the main task categories from the following, and the directions under each category can be diversified:
1. Classification Task \\
- Sentiment Analysis \\
- Text Classification \\
- Toxic Content Detection \\
- Empathy Detection \\
- Stereotype Detection\\
- Social Norm Judgment\\

2. Information Extraction\\
- Named Entity Recognition \\
- Keyphrase Annotation \\
- Coreference Resolution \\
- Entity Relationship Classification\\

3. Text Generation
- Story Creation\\
- Poetry Generation\\
- Recipe Generation\\
- Outline Generation\\
- Text Expansion/Compression\\
- Title Generation\\
- Data Description Generation\\
- Text Rewriting/Simplification\\

4. Dialogue Systems\\
- Dialogue Generation\\
- Intent Recognition\\
- Question Generation/Rewriting\\
- Dialogue State Tracking\\
- Role-playing Dialogue\\

5. Reasoning and Logic\\
- Common Sense QA\\
- Multi-hop QA\\
- Critical Thinking Judgment\\
- Mathematical Reasoning\\
- Theory of Mind Reasoning\\

6. Language Style\\
- Style Transfer\\
- Language Detection\\
- Sarcasm Detection\\
- Spelling/Punctuation Error Detection\\

7. Evaluation and Verification\\
- Text Quality Evaluation\\
- Fact-checking\\
- Answer Verification\\
- Uncertainty Judgment\\

8. Programming-related\\
- Code Generation/Debugging\\
- Code Explanation\\
- Code Translation\\
\begin{verbatim}
Input Total Task:
{input_text}

Input Sub-tasks:
{sub_instruction}
\end{verbatim}
6. First, output the modified comprehensive task content (if no modifications, output the original content) after -output-, including the modified INSTRUCTION and SUB\_INSTRUCTION\_x, where SUB\_INSTRUCTION\_x is formatted as 'sub-task content [task type]'.\\
Finally, place the specific modification analysis process after -explanation-.
\begin{verbatim}
-output-:
INSTRUCTION:
...

SUB_INSTRUCTION_0: 
... [task type]

SUB_INSTRUCTION_1: 
... [task type]

-explanation-:
...
\end{verbatim}

\end{quote}

\subsection{Prompt for constraint-level validation}
\begin{quote}
\tt\small
You are a constraint evaluation assistant.\\
Your task is to determine whether the given constraints can be completed by a large model. Please evaluate according to the following rules:\\

1. **Input**:\\
   - Constraint content: A segment of text describing the task requirements.\\
   - Input dialogue: A segment of the user's conversation with the model.\\

2. **Evaluation Rules**:\\
   - If the input dialogue **lacks the critical information needed to fulfill the constraint** (e.g., the constraint requires extracting person entities, but no person is mentioned in the dialogue), then output ``No''.\\
   - If the constraint **goes beyond the model's capability** (e.g., needs real-time data or external resources), then output ``No''.\\
   - If the input dialogue provides sufficient information and the constraint falls within the model's capability, then output ``Yes''.\\
   - If the model outputs ``No'', minimally modify the constraint content to make it feasible for the model to complete it.\\

3. **Output**:\\
   - If the output is ``No'', provide the modified constraint content to make it feasible for the model.\\
   - If the output is ``Yes'', no modification is needed.\\

4. **Examples**:\\
   - Example 1:\\
     - Instruction content: Extract entities\\
     - Constraint content: Extract person entities from the dialogue.\\
     - Input dialogue: User says, ``Yesterday I went to the park with Xiaoming.''\\
     - Output: Yes\\
   - Example 2:\\
     - Instruction content: Extract entities\\
     - Constraint content: Extract person entities from the dialogue.\\
     - Input dialogue: User says, ``The weather was great yesterday, and I went for a walk in the park.''\\
     - Output: No\\
     - Reason: No person entities in the dialogue\\
     - Modified content: If any person entities are present, extract them.\\
   - Example 3:\\
     - Instruction content: Generate text\\
     - Constraint content: Generate a 100-word text describing the summer scenery, using at least 3 similes.\\
     - Input dialogue: User says, ``Please write a passage about summer.''\\
     - Output: Yes\\
   - Example 4:\\
     - Instruction content: Generate text\\
     - Constraint content: Generate a 100-word text describing the summer scenery, and cite at least 2 academic papers.\\
     - Input dialogue: User says, ``Please write a passage about summer.''\\
     - Output: No\\
     - Reason: Unable to cite academic papers\\
     - Modified constraint: Generate a 100-word text describing the summer scenery, using at least 3 similes.\\

5. **Task**:
\begin{verbatim}
   - Instruction content: {instruction}
   - Constraint content: {constraint}
   - Input dialogue: {input}
   - Output:
   - Reason:
   - Modified constraint:
\end{verbatim}
\end{quote}

\subsection{Prompt for training process}
\begin{quote}
\tt\small
You are now an AI assistant responsible for generating answers to specified tasks. You need to generate answers following these requirements:

1. Strictly generate answers based on the given input material and corresponding sub\_instruction.\\
2. Generate answers for each sub\_instruction, ensuring consistency among answers to different sub\_instructions.\\
3. Follow the constraints of each sub\_instruction strictly to generate answers.\\
4. First, think through each sub-task in detail using analytical skills to deeply understand the issues, and then provide answers. The thought process for each sub-task should be detailed between start\_think and end\_think, and the answer should be fully presented between start\_answer and end\_answer.\\
5. The thought process and answer for each sub\_instruction should be placed between start\_sub\_instruction\_x and end\_sub\_instruction\_x, where sub\_instruction\_x is the specific identifier for the sub-task. Ensure there are no extra spaces, quotes, or symbols before and after these markers.\\

Notes:
1. Strictly adhere to the constraints.\\
2. Ensure the quality of answers.\\
3. Do not output the input content.\\
4. The format is as follows: 

\begin{verbatim}
start_sub_instruction_0
start_think 
Deeply analyze this sub-task, ...
end_think
start_answer
Based on the above analysis, the detailed 
answer to sub-task 0 is ...
end_answer
end_sub_instruction_0

start_sub_instruction_1
start_think 
In this sub-task, consider various 
factors, ...
end_think
start_answer
Based on the above analysis, the answer 
to sub-task 1 is ...
end_answer
end_sub_instruction_1
...
\end{verbatim}

Referring to the above format and generation requirements, please think through and generate specific answers for the following task:
\begin{verbatim}
--input--:
{input_text}
--output--:
\end{verbatim}

\end{quote}

\end{document}